\documentclass{article}
\usepackage{spconf,amsmath,graphicx,hyperref}
\usepackage{booktabs}
\usepackage{amsmath,amssymb}

\title{MotionSpec: Spectral Trajectory Supervision for Motion-Consistent Video Generation}

\name{
Ziqi Ni$^{1}$,
Rui Li$^{2}$,
Shiqi Jiang$^{3}$,
Wei Zhou$^{1,\star}$\thanks{
$\star$ Corresponding author: ZhouW26@cardiff.ac.uk
}
}

\address{
$^{1}$Cardiff University\quad
$^{2}$University of Science and Technology of China \\
$^{3}$East China Normal University
}
\begin{document}
\ninept
\maketitle
\begin{abstract}
Recent advances in text-to-video generation have enabled high-fidelity visual synthesis, yet realistic motion remains challenging. Generated videos may exhibit temporal discontinuities, inconsistent action progression, and structural distortions during complex movements. Even when individual frames appear realistic, the underlying motion may evolve in inconsistent or implausible ways. Standard generative objectives provide limited motion-specific supervision, leaving motion evolution insufficiently constrained. In this paper, we propose MotionSpec, a motion supervision framework centered on Spectral Trajectory Consistency (STC). STC constructs dense anchor-relative motion trajectories and transforms them into motion spectral volumes via a temporal Fourier transform. By aligning the spectral amplitude and phase of predicted and target trajectories, STC constrains both motion strength across temporal frequencies and the temporal organization of motion. To complement this trajectory-level supervision, we introduce Local Flow Consistency (LFC), which aligns consecutive-frame optical flow between predicted and target videos to stabilize local motion transitions. Experiments demonstrate that MotionSpec consistently improves motion consistency, temporal coherence, and plausibility while preserving visual fidelity. We will make our code publicly available.
\end{abstract}
\section{Introduction}
\label{sec:introduction}
Recent advances in text-to-video (T2V) generation have enabled high-fidelity and text-aligned video synthesis with scalable diffusion and flow-matching architectures~\cite{wan,cogvideox,opensora,dit,kong2024hunyuanvideo}. However, strong visual quality does not necessarily imply realistic motion. Generated videos may still exhibit temporal discontinuities, inconsistent action progression, and structural distortions during complex movements~\cite{videojam}. These failures concern not only motion magnitude, but also how motion evolves over time.
A key limitation is the lack of explicit motion supervision in conventional objectives. Standard denoising and flow-matching losses learn appearance and temporal structure jointly through noise or velocity prediction, without directly constraining motion correspondences. Recent methods improve motion through joint appearance-motion learning~\cite{videojam,jeong2025track4gen}, decomposed conditioning~\cite{demo}, optical-flow supervision~\cite{flowloss}, motion-aware loss reweighting~\cite{motif,ltd,zhao2024motiondirector}, or flow-guided prompt optimization~\cite{motionprompt}. Spectral regularization has also been explored to improve motion in video generation~\cite{physicsmotionloss,freeinit,sma,lei2025animateanything,ni2025freak}.
However, these approaches do not explicitly align the
temporal spectra of dense motion trajectories.

\begin{figure}[t]
    \centering
    \includegraphics[width=\linewidth]{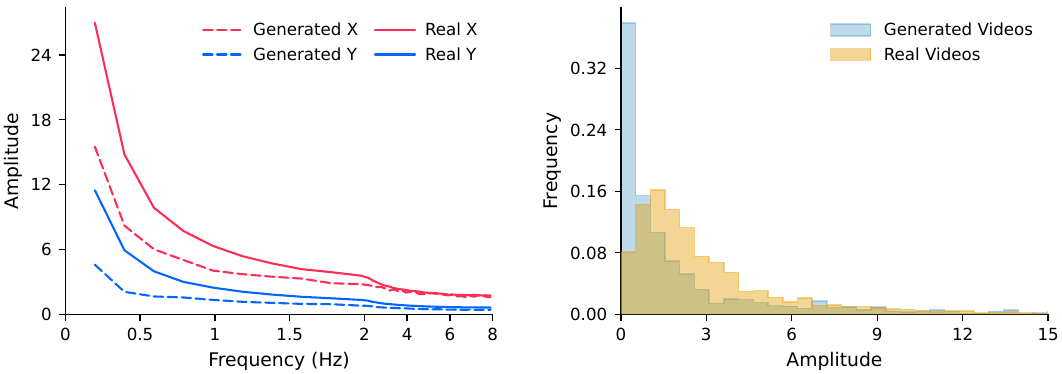}
    \caption{Spectral volume statistics computed from 500 generated-real video pairs using 81 frames sampled from the beginning of each video at 16 fps.
    \textbf{Left}: mean amplitudes of non-zero-frequency components of horizontal and vertical first-frame-anchored motion trajectories.
    \textbf{Right}: normalized distribution of per-video root-mean-squar (RMS)
    amplitudes averaged over 40 frequency bins.
    }
    \label{fig:motiv}
\end{figure}

To characterize motion over multiple frames, we analyze the temporal spectra of dense motion trajectories. As shown in Fig.~\ref{fig:motiv}, generated videos in our evaluation exhibit lower mean spectral amplitudes at nonzero temporal frequencies than real videos, with the largest absolute gaps at low temporal frequencies. The per-video amplitude statistics are also more concentrated near zero, whereas real videos exhibit a broader distribution.
Together, these observations indicate a tendency toward weaker trajectory variation in generated videos.
This motivates explicit supervision of motion trajectories across multiple frames through their temporal-frequency representations.

Motivated by this observation, we propose MotionSpec, a motion supervision framework centered on Spectral Trajectory Consistency (STC). We construct dense anchor-relative displacement trajectories and transform them into motion spectral volumes (MSV)~\cite{generativeimagedynamics}. STC explicitly aligns the trajectory spectra of generated videos with those of real videos through amplitude and phase supervision, constraining both the distribution of motion energy across temporal frequencies and the temporal organization of motion. To complement this trajectory-level spectral supervision, we further introduce Local Flow Consistency (LFC), which aligns adjacent-frame optical flow between predicted and target videos to stabilize local motion transitions. Together, these objectives provide explicit supervision for motion evolution during video generation.
\begin{figure*}
    \centering
    \includegraphics[width=\linewidth]{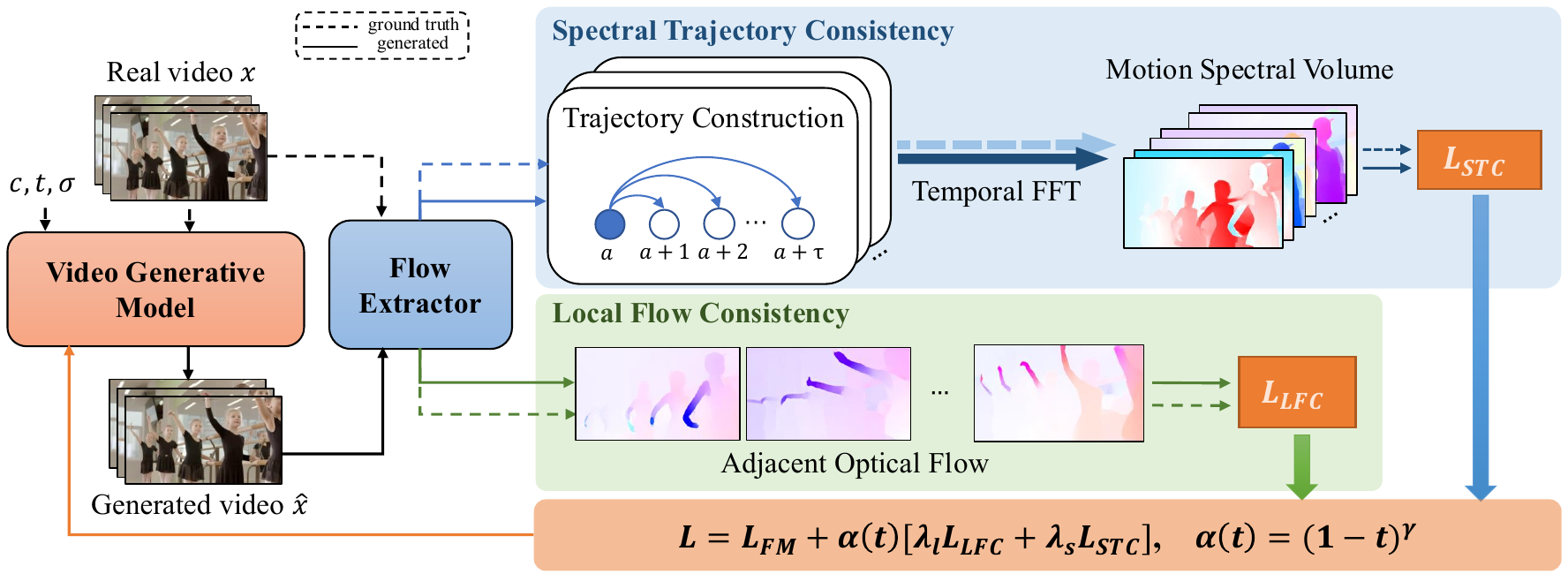}
    \caption{Overview of the pipeline. At each training step, the video generative model takes the prompt $c$, timestep $t$, noise level $\sigma$, and real data as inputs. After obtaining the clean latent and decoding it into a video, both the generated and ground-truth videos are fed into an optical flow estimator to compute the Local Flow Consistency loss and the Spectral Trajectory Consistency loss. These objectives are jointly optimized with the flow matching loss.}
    \label{fig:method}
\end{figure*}

Our main contributions are summarized as follows:
\begin{itemize}
\item We introduce Spectral Trajectory Consistency (STC), a motion supervision objective that aligns the temporal spectra of dense motion trajectories between generated and real videos.
\item We develop MotionSpec, which incorporates STC into video generation and complements it with LFC to stabilize adjacent-frame motion transitions.
\item Experiments show consistent improvements in motion quality, temporal coherence, and structural stability while preserving visual appearance.
\end{itemize}

\section{Method}
\label{sec:method}
\subsection{Overview}
\textbf{Preliminaries.} Let $z_0$ denote the latent representation of a $T$-frame training
video $\mathbf{x}$ with text condition $c$.
In rectified flow~\cite{rf}, $z_t=(1-t)z_0+t z_1$, where $z_1\sim\mathcal{N}(0,I)$ and
$t\in[0,1]$.
The model predicts velocity $v_\theta(z_t,t,c)$ and obtains the
clean-video estimate
$\hat{\mathbf{x}}=\mathcal{D}(z_t-t\,v_\theta(z_t,t,c))$,
where $\mathcal{D}$ is the video decoder. The flow-matching objective is:
\begin{equation}
    \mathcal{L}_{\mathrm{FM}}
    =
    \mathbb{E}_{z_0,z_1,t,c}
    \left[
        \left\|
            v_\theta(z_t,t,c)-(z_1-z_0)
        \right\|_2^2
    \right].
    \label{eq:fm}
\end{equation}

Fig.~\ref{fig:method} illustrates the overall MotionSpec framework.
The predicted and ground-truth videos are passed through a frozen optical-flow estimator to extract motion representations. Our main objective, Spectral Trajectory Consistency (STC), constructs anchor-relative motion trajectories and supervises their temporal spectra to constrain longer-range motion evolution. We additionally employ Local Flow Consistency (LFC) to align adjacent-frame optical flow and stabilize local motion transitions. Both objectives are jointly optimized with the flow-matching loss.

\subsection{Local Flow Consistency}
\label{sec:short}
We enforce local flow consistency (LFC) by aligning adjacent-frame optical flow between the predicted and target videos. For motion extraction, we use a frozen, differentiable optical-flow
estimator $\mathcal{O}$. Let $F_i=\mathcal{O}(x_i,x_{i+1})$ for $i=1,\ldots,T-1$. 
To emphasize moving regions, we construct a soft mask from the target flow magnitude:
\begin{equation}
    M_i(p)
    =
    \operatorname{clip}
    \left(
        \frac{\|F_i(p)\|_2}{\tau_m},\,0,\,1
    \right),
    \label{eq:motion_mask}
\end{equation}
where $p$ indexes spatial locations and $\tau_m>0$ is a motion threshold. Using the Charbonnier penalty
$\rho(r)=\sqrt{\|r\|_2^2+\delta^2}$, we define:
\begin{equation}
\begin{split}
    \mathcal{L}_{\mathrm{LFC}}
    &=
    \frac{1}{T-1}
    \sum_{i=1}^{T-1}
    \frac{1}{\sum_p M_i(p)+\varepsilon}
    \\
    &\quad\times
    \sum_p M_i(p)\,
    \rho\big(F_i(p)-\hat{F}_i(p)\big),
\end{split}
    \label{eq:lshort}
\end{equation}
where $\delta>0$ controls the smoothness of the penalty near zero
and $\varepsilon>0$ ensures numerical stability.

\subsection{Spectral Trajectory Consistency}
\label{sec:long}
Adjacent-frame flow does not explicitly constrain motion evolution over a longer temporal horizon. A generated video may therefore contain plausible local transitions while exhibiting unrealistic motion decay, trajectory drift, or inconsistent long-term dynamics.
To capture such patterns, we construct anchor-relative motion
trajectories. 
Given a temporal anchor $a$, the target displacement from
the anchor frame to a future frame is:
\begin{equation}
    G_{a,\tau}
    =
    \mathcal{O}(x_a,x_{a+\tau}),
    \label{eq:target_long_flow}
\end{equation}
where $\tau=1,\ldots,L$ and $L$ is the trajectory horizon. 
We use a set of sliding temporal anchors that start from the first frame
and advance with stride $s$. Only anchors satisfying $a+L\leq T$ are
retained. This allows the spectral objective to characterize multiple
temporal segments within the same video.

For each spatial location $p$ and flow component
$d\in\{u,v\}$, we apply a one-dimensional Fourier
transform (1D-FFT) along its temporal dimension:
\begin{equation}
\begin{split}
    S_{a,k}(p,d)
    &=
    \sum_{\tau=1}^{L}
    G_{a,\tau}(p,d)
    \exp\left(
        -j\frac{2\pi k(\tau-1)}{L}
    \right).
\end{split}
    \label{eq:target_msv}
\end{equation}

The resulting complex-valued tensors form the motion spectral volumes. The predicted $\hat{S}_{a,k}$ is defined similarly. We discard the zero-frequency (DC) component and retain the first $K$ positive-frequency bins. And we define the logarithmic spectral amplitude as:
\begin{equation}
    A_{a,k}(p,d)
    =
    \log\left(
        1+|S_{a,k}(p,d)|
    \right).
    \label{eq:target_amplitude}
\end{equation}

Because phase becomes unreliable when the target spectral amplitude is
close to zero, we construct an amplitude-dependent phase confidence:
\begin{equation}
\begin{split}
    Q_{a,k}(p,d)
    &=
    \operatorname{clip}
    \left(
        \frac{|S_{a,k}(p,d)|}{\tau_\phi},
        0,1
    \right).
\end{split}
    \label{eq:phase_confidence}
\end{equation}

For two complex coefficients $s$ and $\hat{s}$, we define their
circular phase distance as:
\begin{equation}
\begin{split}
    d_\phi(s,\hat{s})
    &=
    1-
    \frac{
        \operatorname{Re}
        \left(s\hat{s}^{*}\right)
    }{
        |s|\,|\hat{s}|+\varepsilon
    }.
\end{split}
    \label{eq:phase_distance}
\end{equation}
This formulation avoids the discontinuity caused by directly
subtracting phase angles around the $2\pi$ boundary.
The spectral amplitude loss is:
\begin{equation}
\begin{split}
    \mathcal{L}_{\mathrm{amp}}
    &=
    \frac{1}{|\mathcal{A}|}
    \sum_{a\in\mathcal{A}}
    \sum_{k=1}^{K}
    w_k
    \left\langle
        |A_{a,k}-\hat{A}_{a,k}|
    \right\rangle_{p,d},
\end{split}
    \label{eq:lamp}
\end{equation}
where $\langle\cdot\rangle_{p,d}$ denotes averaging over spatial
locations and flow components. The non-negative frequency weights
$w_k$ are normalized to sum to one.
The phase consistency loss is:
\begin{equation}
\begin{split}
    \mathcal{L}_{\mathrm{phase}}
    &=
    \frac{1}{|\mathcal{A}|}
    \sum_{a\in\mathcal{A}}
    \sum_{k=1}^{K}
    w_k
    \left\langle
        Q_{a,k}\,
        d_\phi
        \left(
            S_{a,k},
            \hat{S}_{a,k}
        \right)
    \right\rangle_{p,d}.
\end{split}
    \label{eq:lphase}
\end{equation}

And the complete spectral trajectory consistency objective is:
\begin{equation}
    \mathcal{L}_{\mathrm{STC}}
    =
    \mathcal{L}_{\mathrm{amp}}
    +
    \lambda_\phi
    \mathcal{L}_{\mathrm{phase}},
    \label{eq:lstc}
\end{equation}
where $\lambda_\phi$ balances amplitude and phase supervision. The
amplitude term aligns motion energy across temporal frequencies, while
the phase term constrains the temporal organization of long-range
motion trajectories.

\subsection{Training Objective}
\label{sec:objective}
Optical flow estimated from the predicted video becomes less reliable at highly noisy timesteps. We therefore introduce the timestep-dependent weight:
\begin{equation}
    \alpha(t)
    =
    (1-t)^\gamma.
    \label{eq:timestep_weight}
\end{equation}
Under our convention, the clean-data endpoint corresponds to $t=0$,
whereas the pure-noise endpoint corresponds to $t=1$.
The final training objective is:
\begin{equation}
\begin{split}
    \mathcal{L}
    &=
    \mathcal{L}_{\mathrm{FM}}
    +
    \alpha(t)
    \Big(
        \lambda_l
        \mathcal{L}_{\mathrm{LFC}}
        +
        \lambda_s
        \mathcal{L}_{\mathrm{STC}}
    \Big),
\end{split}
    \label{eq:total}
\end{equation}
where $\lambda_l$ and $\lambda_s$ control the strengths of short-term
flow supervision and long-term spectral trajectory supervision,
respectively.

\section{Experiments}
\label{sec:experiments}
\subsection{Experimental Setup}
\label{sec:exp_setup}
\noindent\textbf{Dataset.}
We filtered 10K training videos from OpenVid-1M~\cite{openvid}
based on motion magnitude and prompt content, covering human,
object, and complex motion.

\noindent\textbf{Backbone.}
We choose Wan2.1-T2V-1.3B\cite{wan} as our backbone for its representative architecture, and manageable model size, which facilitates single-GPU fine-tuning and multiple ablation studies.

\noindent\textbf{Evaluation Metrics.}
We primarily adopt VMBench~\cite{vmbench} to assess motion quality
from multiple complementary aspects.
We additionally report appearance and motion scores following
VideoJAM~\cite{videojam}, computed by aggregating the corresponding
VBench~\cite{vbench} metrics.

\noindent\textbf{Implementation Details.}
We train on video clips of size $256 \times 256 \times 81$ and set $\gamma=2$ and $\lambda_l=\lambda_s=\lambda_\phi=1$. For STC, we use $L=60$ and $s=10$ to cover the early, middle, and late portions of each video while maintaining a long temporal span for trajectory supervision. We retain the first $K=16$ frequency bins to focus supervision on lower temporal frequencies. 
All experiments are conducted on a single NVIDIA H200 GPU.

\subsection{Quantitative Results}
\label{sec:quantitative}
\begin{table}[t]
    \centering
    \caption{Quantitative comparison. FM-only denotes fine-tuning with the flow-matching objective alone. $+$ LFC and $+$ STC add the respective objective to FM-only, while Ours incorporates both.}
    \label{tab:main}
    \setlength{\tabcolsep}{2.2pt}
    \resizebox{\linewidth}{!}{
    \begin{tabular}{lcccccccc}
        \toprule
        & \multicolumn{2}{c}{VideoJAM-Bench}
        & \multicolumn{6}{c}{VMBench} \\
        \cmidrule(lr){2-3}
        \cmidrule(lr){4-9}
        Method
        & App.$\uparrow$
        & Mot.$\uparrow$
        & CAS$\uparrow$
        & MSS$\uparrow$
        & OIS$\uparrow$
        & PAS$\uparrow$
        & TCS$\uparrow$
        & Avg.$\uparrow$ \\
        \midrule

        Wan2.1
        & 81.25 & 85.3
        & 54.13 & 85.12 & 48.34 & 23.17 & 97.09 & 61.57 \\
        FM-Only
        & 81.95 & 87.43
        & 53.83 & 87.01 & 50.25 & 22.91 & 96.88 & 62.18 \\
        $+$ LFC
        & 81.72 & 90.13
        & 53.97 & 89.91 & 54.34 & \textbf{24.27}& 97.97 & 64.09 \\
        $+$ STC
        & 82.15 & 92.45
        & 55.77 & 89.72 & 50.48 & 23.65 & 98.73 & 63.67 \\
        \textbf{Ours}
        & \textbf{82.2} & \textbf{93.61}
        & \textbf{56.27} & \textbf{89.99} & \textbf{56.53} & 23.74 & \textbf{98.82} & \textbf{65.07} \\

        \bottomrule
    \end{tabular}
    }
\end{table}
Table~\ref{tab:main} presents the quantitative results. To control for the effect of curated dataset, we include an FM-only baseline that uses the same training configuration but optimizes only the flow-matching objective. FM-only increases the average VMBench score from 61.57 to 62.18, indicating that fine-tuning alone o n the dataset contributes to the improvement. Both LFC and STC provide further gains, and their combination achieves the best overall performance. 

The two objectives exhibit complementary performance profiles. LFC achieves a higher object integrity score than STC, whereas STC achieves higher temporal coherence  and VideoJAM-Bench motion scores. Combining both objectives further improves OIS to 56.53 and TCS to 98.82, outperforming either component alone on these dimensions. 
Our full model also improves the appearance score from 81.95 to 82.20 compared with FM-only. Its PAS increases from 22.91 to 23.74, indicating that the gains in motion quality are accompanied by an increase in perceptible motion amplitude. While LFC alone achieves the highest PAS, the full model obtains the highest average VMBench score, with improvements over FM-only across all five dimensions.

\begin{figure*}
    \centering
\includegraphics[width=\linewidth]{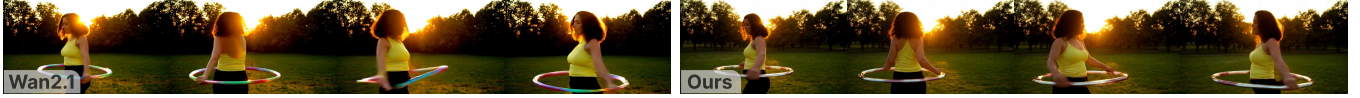}
    \caption{\textbf{Qualitative Results.}
On the left are the results of Wan2.1, and on the right are the results of ours. It can be observed that in our results, the motion patterns are more realistic and the motion performance is significantly better.}
    \label{fig:quali}
\end{figure*}

\subsection{Qualitative Results}
\label{sec:qualitative}
Figure~\ref{fig:quali} presents qualitative comparisons between Wan2.1 and our method. In the first row, the boy generated by Wan2.1 appears to cycle backward toward the camera, producing an unnatural motion pattern. In contrast, our method generates a boy facing the camera and riding naturally toward it, showing better consistency between body orientation and the direction of motion.
In the second and third rows, Wan2.1 exhibits limb distortions and blurred body regions during running, whereas our method produces more coherent body structures and clearer motion across frames. These examples highlight the difficulty of preserving structural consistency under fast and complex human motion. In the fourth row, which depicts a girl hula-hooping, Wan2.1 shows inconsistent clothing appearance and deformations of the hoop during the motion. Our method maintains more stable clothing appearance and better preserves the hoop's shape throughout the sequence.

These observations are consistent with our motion objectives. 
Motion-masked short-term flow supervision constrains local motion direction and magnitude in moving regions, while spectral trajectory consistency aligns the amplitude and phase spectra of anchor-relative motion trajectories to constrain longer-term motion evolution and temporal organization. Together, these objectives encourage coherent local transitions and more consistent motion dynamics, which may contribute to the improved structural stability and visual consistency observed in these examples.

\subsection{Ablation Studies}
\label{sec:frequency}
Table~\ref{tab:main} has already demonstrated the individual and complementary effects of LFC and STC. We further analyze the internal components of STC, namely the spectral amplitude loss and the phase consistency loss. To isolate their contributions, all variants in this study are trained with LFC enabled, and differ only in the spectral supervision applied on top of it.
As shown in Table~\ref{tab:stc_ablation}, introducing either amplitude or phase supervision improves the motion-related metrics over the LFC-only baseline. The amplitude term mainly improves motion smoothness and structural stability by aligning the distribution of motion energy across temporal frequencies. In contrast, the phase term provides a larger gain in temporal coherence, indicating that phase information is important for preserving the temporal organization of long-range trajectories. Combining both terms yields the best overall performance.

\begin{table}[t]
    \centering
    \caption{Ablation study on the components of spectral trajectory consistency.}
    \label{tab:stc_ablation}
    \small
    \setlength{\tabcolsep}{2.8pt}
    \begin{tabular}{lccccccc}
        \toprule
        Variant
        & Mot.$\uparrow$
        & CAS$\uparrow$
        & MSS$\uparrow$
        & OIS$\uparrow$
        & PAS$\uparrow$
        & TCS$\uparrow$
        & Avg.$\uparrow$ \\
        \midrule
        LFC
        & 90.13
        & 53.97
        & 89.91
        & 54.34
        & \textbf{24.27}
        & 97.97
        & 64.09 \\
        $+$ Amplitude
        & 91.84
        & 54.82
        & \textbf{90.21}
        & 55.18
        & 24.08
        & 98.31
        & 64.52 \\
        $+$ Phase
        & 91.36
        & 55.21
        & 90.04
        & 54.87
        & 23.69
        & 98.54
        & 64.47 \\
        \textbf{Full(Ours)}
        & \textbf{93.61}
        & \textbf{56.27}
        & 89.99
        & \textbf{56.53}
        & 23.74
        & \textbf{98.82}
        & \textbf{65.07} \\
        \bottomrule
    \end{tabular}
\end{table}

\begin{table}[t]
\centering
\caption{Ablation of time-domain and spectral trajectory supervision.}
\label{tab:traj_domain}
\resizebox{\columnwidth}{!}{
\begin{tabular}{lccccccc}
\toprule
Variant & Mot.$\uparrow$ & CAS$\uparrow$ & MSS$\uparrow$ & OIS$\uparrow$ & PAS$\uparrow$ & TCS$\uparrow$ & Avg.$\uparrow$ \\
\midrule
LFC 
& 90.13 & 53.97 & 89.91 & 54.34 &\textbf{24.27} & 97.97 & 64.09 \\

+ Time-domain 
& 92.05 & 55.10 & \textbf{90.12} & 55.46 & 23.88 & 98.42 & 64.60 \\

\textbf{+ STC(Ours) }
& \textbf{93.61} & \textbf{56.27} & 89.99 & \textbf{56.53} & 23.74 & \textbf{98.82} & \textbf{65.07} \\
\bottomrule
\end{tabular}
}
\end{table}

To determine whether the gain of STC merely come from long-range trajectory supervision, we compare it with a direct time-domain loss. As shown in Table~\ref{tab:traj_domain}, time-domain supervision improves the motion and average scores over LFC alone, confirming the benefit of constraining long-range motion. Nevertheless, STC achieves higher scores, increasing the average score to 65.07. This suggests that STC benefits not only from a longer supervision horizon. By decomposing trajectories into temporal frequency components, STC separates motion energy from temporal organization through amplitude and phase supervision, providing a more structured constraint on motion evolution across temporal scales. In contrast, point-wise temporal matching penalizes trajectory discrepancies at individual temporal offsets without explicitly distinguishing slowly varying trends from faster temporal variations.

\section{Conclusion}
We present MotionSpec, a motion supervision framework for motion-consistent video generation. STC constrains the frequency structure and temporal organization of motion trajectories, providing explicit supervision for motion evolution beyond adjacent frames. We complement STC with local flow consistency to stabilize local motion transitions in moving regions. Together, these objectives improve motion smoothness, structural stability, and temporal coherence without sacrificing visual fidelity. Experiments demonstrate consistent improvements across motion-oriented evaluation metrics.

\vfill\pagebreak

\bibliographystyle{IEEEbib}
\bibliography{strings,refs}

\end{document}